# Constructing Locally Dense Point Clouds Using OpenSfM and ORB-SLAM2

Fouad Amer
Civil Engineering Department
University of Illinois at Urbana-Champaign
E-mail: amer.fouad@gmail.com

Wilfredo Torres
Civil Engineering Department
University of Illinois at Urbana-Champaign
E-mail: wtorresc5@gmail.com

Zixu Zhao
Electrical and Computer Engineering Department
University of Illinois at Urbana-Champaign
E-mail: zzhao39@illinois.edu

Siwei Tang
Mechanical Engineering Department
University of Illinois at Urbana-Champaign
E-mail: siweit2@illinois.edu

**Abstract**

*This paper aims at finding a method to register two different point clouds constructed by ORB-SLAM2 and OpenSfM. To do this, we post some tags with unique textures in the scene and take videos and photos of that area. Then we take short videos of only the tags to extract their features. By matching the ORB feature of the tags with their corresponding features in the scene, it is then possible to localize the position of these tags both in point clouds constructed by ORB-SLAM2 and OpenSfM. Thus, the best transformation matrix between two point clouds can be calculated, and the two point clouds can be aligned.*

## 1. Introduction

ORB-SLAM2 and OpenSfM are two methods of constructing point clouds in a certain area. ORB-SLAM2 can construct sparse map with high speed, while OpenSfM can generate high-quality dense point cloud with relatively lower speed. In the context of building construction work, the subcontractor may only care about the indoor construction of some certain area. So, we want to generate a high-quality point cloud of that area. However, using OpenSfM to get a high-quality point cloud of the whole indoor environment could be extremely time consuming. Therefore, we come up with a method to align two point clouds which are generated by ORB-SLAM2 and OpenSfM respectively. In that case, we can combine the localization and speed of ORB-SLAM2 with the quality, accuracy and density of 3D construction that can be obtained using OpenSfM. Once we find the alignment of two kinds of point clouds, we can use ORB-SLAM2 to construct 3D point cloud of the whole indoor environment and use OpenSfM to generate point cloud of the certain area we are interested in. Then we align these two point clouds together to get the sparse map of the whole indoor environment and dense map of the important area. Finally, we can achieve our goal and save a lot of time running code.

## 2. Software Installation and Data Collection

[1] [2] ORB-SLAM2 is a SLAM library that computes the camera trajectory and a 3D reconstruction of the scene in real-time. It's also capable of localizing the camera and detecting loops. To reconstruct the experiment results in this project, other dependencies like Pangolin, OpenCV, and Eigen3 are needed.

In this project, SLAM is used for calculating the global sparse map of the building. Since the goal is to register two different point clouds, precise geological information of each ORB feature in the point cloud is desired. Therefore, the trajectory of the video used for SLAM must form a closed loop to obtain accurate coordinates of all the points.

[3] OpenSfM is a Structure from Motion library written in Python on top of OpenCV. Other dependencies for the software are OpenCV, OpenGV, Ceres Solver, Boost Python, Numpy, Scipy, Networkx, PyYAML, and exifread. SFM is used for reconstructing the high-quality point cloud for some specific locations, so we only take images of the interested area. The specific locations where a dense reconstruction is required must be on the trajectory of the SLAM video.

Tags are used for this project to obtain the geological location of their interest points in two different point clouds. This can help us register the point clouds by calculating the affine transformation matrix. The tags used for this project are images with unique textures, and they are used as locations that have special features that can be easily searched and found. These tags are posted at the location where the alignment is needed. A minimum of 4 tags is needed to calculate the transformation matrix.

## 3. Generating Point Clouds

### 3.1. ORB-SLAM2

Extract frames from the video. A proper frame rate should be high enough to guarantee the continuousness of the camera trajectory and low enough that so that the algorithm won't run too long. Several frame rates are tested in the experiment, and it's found that 29 meets the above conditions.

Run the installed ORB-SLAM2 on the extracted frames, and save the ORB [4] descriptors, their point coordinates and the frame trajectory. Notice that the ORB descriptors are binary but the saved ones are integers, so they should be converted to binary number in two's complement form.

Run ORB-SLAM2 on short videos of the tags to extract the ORB feature and extract the ORB features.

### 3.2. OpenSfM

Run OpenSfM on the available photos or videos. In OpenSfM we activate the depth maps reconstruction option to get densified high-quality point cloud. The required outputs are the tracks file, the json reconstruction file, the saved descriptors, and the merge.ply file. Then, run OpenSfM over pictures of the tags to extract their HaHOG (the combination of Hessian Affine feature point detector and HOG descriptor) features, which uses SIFT [5] descriptor.

At this stage, the data is available for localization and alignment.

### 3.3. Localizing the Tags in ORB-SLAM2 Point Cloud

In the point cloud generated by ORB-SLAM2, we firstly transform the extracted ORB descriptor into their binary representation. Then Localize (find the coordinates) tags within the point cloud by localizing points associated with each tag using descriptor matching described in 3.5. We implement the function "orb feature matcher" in MATLAB in this step.

### 3.4. Localizing the Tags in OpenSfM Point Cloud

In the SFM point cloud, we need to find the descriptors associated with all the points. From the json reconstruction file, we can track the "general id" of every point and its 3d coordinates. Then from the tracks file, for each image that contains a specific set of 3D points, we can find the specific id for each point contained in that image. After that, from the folders containing the descriptors of all the points contained in each image, we can match all the point descriptors with their ids based on the descriptors previously found in an image and their associated feature ids. Finally, we localize tags within the point cloud by localizing points associated with each tag using descriptor matching described in 3.5.

### 3.5. Descriptor Matching

Before alignment, we need to find the coordinates of each tag in two point clouds. To get the coordinate of each tag in ORB-SLAM2 point cloud, for each point 'a' in an image we find two descriptors 'b' and 'c' in the point cloud that have minimum and second minimum Hamming distance (since ORB features are binary) with the descriptor of point 'a'. If the ratio of these two distances is less than 0.7 and the minimum distance is less than 0.35(we got this threshold based on several trials), we can save the point 'b' as a potential match point. After we matched every point in an image, we can get a set of potential match points distributed in the whole area. We know that the real match points of a single tag should be geometrically close to each other in the point cloud because of the geometric constraints of a tag. So, we search for 4 or more points among those potential match points that are geometrically close to each other and save the centroid of these matching points as the coordinate of that tag.

The matching process for OpenSfM is roughly the same. The only difference is that the distance used this time is L2 distance (for HaHog feature) and the threshold for minimum distance is 1(this is obtained by several trials).

## 4. Alignment

After obtaining the coordinate of each tag in two point clouds. We can calculate the transformation matrix between two point clouds. Note that the DOF of the transformation matrix is 7(rotation, translation and scale). We firstly use DLT to transform X' = HX to AH = 0(H is a 4x4 matrix where the last row is [0 0 0 1]). Then we solve it using SVD. Finally, we can get a H that consists the least square solution.

## 5. Results

The data we used for experiment was collected in Newmark Civil Engineering Laboratory. The point clouds generated by ORB-SLAM2 and OpenSfM were aligned together using the above described algorithm and visualized using CloudCompare. The visualized alignment result is shown below:

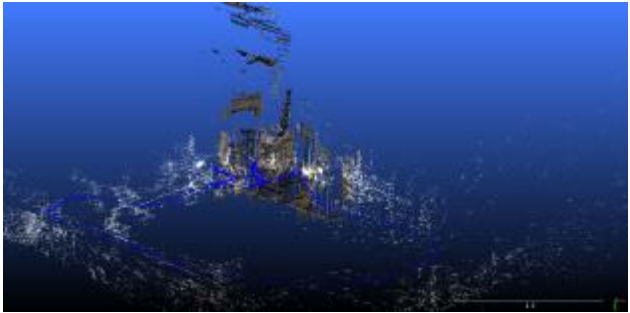

Figure 1: Aligned Point Clouds 1

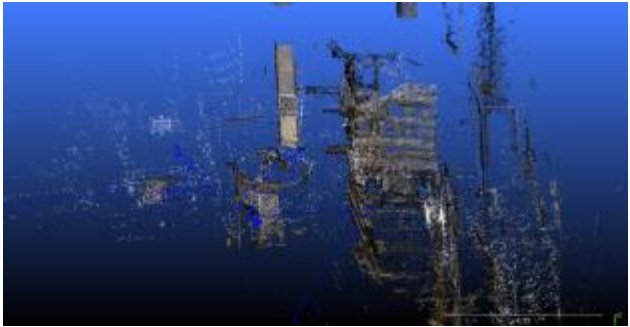

Figure 2: Aligned Point Clouds 2

6. Individual Contribution

Collecting data for both ORB-SLAM2 and openSFM. (*Siwei Tang*)

Creating sparse map of the environment using ORB-SLAM2. (*Zixu Zhao*)

Reconstructing a point cloud using OpenSfM and final registration of the reconstructed point clouds. (*Wilfredo Torres*)

Processing initial output data, feature matching, and finding individual tags within the two point clouds (*Fouad Amer*)

References

[1] Raulmur. "Raulmur/ORB_SLAM2." *GitHub*. H.P., 20 Jan. 2017. Web. 11 May 2017.
[2] Raúl Mur-Artal and Juan D. Tardós. ORB-SLAM2: an Open-Source SLAM System for Monocular, Stereo and RGB-D Cameras. *ArXiv preprint arXiv:1610.06475*.
[3] Mapillary. "Mapillary/OpenSfM." *GitHub*. N.p., 19 Apr. 2017. Web. 11 May 2017.
[4] Rublee, Ethan, Vincent Rabaud, Kurt Konolige, and Gary Bradski. "ORB: An efficient alternative to SIFT or SURF." *2011 International Conference on Computer Vision* (2011): n. pag. Web.
[5] D.G. Lowe, Distinctive image features from scale-invariant keypoints, Int J Comput Vis, 60 (2004) 91-110.